%% file: nips09.tex
\documentclass{article}
\usepackage{nips07submit_e,times}
\usepackage{enumerate}
\usepackage{tabularx}
\usepackage{amsmath}
\usepackage{amsthm}
\usepackage{graphicx}
\usepackage{tikz}
\usepackage{float}
\usepackage[small,bf]{caption}
\usetikzlibrary{arrows}
\usetikzlibrary{shapes.geometric}
\usetikzlibrary{decorations.pathmorphing}

\title{Hierarchies in Dictionary Definition Space}

\author{
Olivier Picard$^1$, \enskip Alexandre Blondin Mass\'e$^2$, \enskip Stevan Harnad$^1$,\\
\textbf{\enskip Odile Marcotte$^3$, \enskip Guillaume Chicoisne$^1$, \enskip Yassine Gargouri$^1$} \\
\And
\rm $^1$ Institut des sciences cognitives \\
Universit\'e du Qu\'ebec \`a Montr\'eal (UQ\`AM)\\
Montr\'eal (Qu\'ebec) Canada H3C 3P8\\
{\footnotesize\texttt{picard.olivier.2@courrier.uqam.ca}},
{\footnotesize\texttt{harnad@uqam.ca}}, \\
{\footnotesize\texttt{chicoisne.guillaume@uqam.ca}},
{\footnotesize\texttt{yassinegargouri@hotmail.com}} \\
\And
\rm $^2$ Laboratoire de combinatoire et d'informatique math\'ematique\\
Universit\'e du Qu\'ebec \`a Montr\'eal (UQ\`AM)\\
Montr\'eal (Qu\'ebec) Canada H3C 3P8\\
{\footnotesize\texttt{alexandre.blondin.masse@gmail.com}} \\
\And
\rm $^3$ Groupe d'\'etudes et de recherche en analyse des d\'ecisions (GERAD) and UQ\`AM \\
HEC Montr\'eal \\
Montr\'eal (Qu\'ebec) Canada H3T 2A7\\
{\footnotesize\texttt{Odile.Marcotte@gerad.ca}}
}
\begin{document}

\maketitle

\newtheorem{definition}{Definition}
\newtheorem{proposition}{Proposition}
\newtheorem*{example}{Example}

\newcommand{\OUTZERO}{\textsc{Out0}}
\newcommand{\SINKS}{\textsc{Sinks}}
\newcommand{\GK}{\textsc{gk}}
\newcommand{\KC}{\textsc{kc}}
\newcommand{\SCC}{\textsc{scc}}
\newcommand{\CIDE}{\textsc{cide}}
\newcommand{\LDOCE}{\textsc{ldoce}}

\pgfkeys{/tikz/i/.style={orange}}
\pgfkeys{/tikz/c/.style={red}}
\pgfkeys{/tikz/aoa/.style={blue}}
\pgfkeys{/tikz/tlf/.style={red!70!blue!40!green!90}}
\pgfkeys{/tikz/bf/.style={red!10!blue!40!green!70}}
\pgfkeys{/tikz/filli/.style={fill=orange}}
\pgfkeys{/tikz/fillc/.style={fill=red}}
\pgfkeys{/tikz/fillaoa/.style={fill=blue}}
\pgfkeys{/tikz/filltlf/.style={fill=red!70!blue!40!green!90}}
\pgfkeys{/tikz/fillbf/.style={fill=red!10!blue!40!green!70}}

\begin{abstract}
A dictionary defines words in terms of other words. Definitions can tell you the meanings of words you don't know, but only if you know the meanings of the defining words. How many words do you need to know (and which ones) in order to be able to learn all the rest from definitions? We reduced dictionaries to their ``grounding kernels" (GKs), about 10\% of the dictionary, from which all the other words could be defined. The GK words turned out to have psycholinguistic correlates: they were learned at an earlier age and more concrete than the rest of the dictionary. But one can compress still more: the GK turns out to have internal structure, with a strongly connected ``kernel core" (KC) and a surrounding layer, from which a hierarchy of definitional distances can be derived, all the way out to the periphery of the full dictionary. These definitional distances, too, are correlated with psycholinguistic variables (age of acquisition, concreteness, imageability, oral and written frequency) and hence perhaps with the ``mental lexicon" in each of our heads.

\textbf{Keywords:} categories, definition, dictionary,  feedback vertex set, graph theory, language learning, lexicography, mental lexicon, semantics, symbol grounding, vocabulary, word meaning
\end{abstract}

\section{Introduction}

A category is a \emph{kind} of thing (object, event, action, trait or state). To categorize is to do the right thing (eat, fight, flee, mate, etc.) with the right kind of thing. All species can acquire categories through trial and error \emph{sensorimotor induction}. We are the only species that can also acquire and transmit categories through \emph{verbal instruction}, by naming and defining them. The words in our dictionaries are almost all the names of categories, followed by their definitions. In principle, all categories can be acquired through verbal definition, but we cannot acquire all of them that way: we have to know the meanings of some of the defining words already, by some other means. This is the ``symbol grounding problem" \cite{harnad} and presumably that other means of acquiring categories is sensorimotor induction. But how many words -- and which ones -- need to be grounded directly through sensorimotor induction in order to allow all the rest to be acquired through verbal definition?

We have been analyzing dictionaries in order to answer this question. By eliminating all the words that can be reached from other words through definition alone, we have been able to reduce the dictionary to its ``grounding kernel" (GK) -- a set of words (about $10\%$) -- out of which all the rest of the words can be reached through definition alone \cite{textgraphs}. The GK has some striking properties: The words in it are learned at a significantly younger age than the rest of the dictionary and are also more concrete \cite{chicoisne}, but if the variance correlated with age is removed, the residual GK words are more abstract than the rest of the dictionary. What is the cause of this polarity shift?

The GK is unique, and sufficient to ground all the rest of the dictionary, but it is not \emph{minimal} -- it is not the smallest set of words from which all the rest can be reached via definition alone. That would be a ``minimum grounding set" (MGS), which is not in general unique; we have not yet been able to compute a MGS, because this problem (equivalent to finding a ``minimum cardinality feedback vertex set'' for a general graph) is NP-complete (i.e. too hard to compute in general). We hope to be able to compute MGSs for our special cases, but meanwhile the GKs of our dictionaries -- Cambridge International Dictionary of English (CIDE) \cite{cide} and Longman Dictionary of Contemporary English (LDOCE) \cite{ldoce} -- already turn out to have more differentiated internal substructure which we begin analyzing further in this article.

In particular two substructures play important roles: the GK itself and a strongly connected subset of the GK that we call the ``Kernel Core" (KC). The GK words that are acquired earlier, and are more concrete than the rest of the dictionary, tend to be in the KC, whereas the GK words uncorrelated with age of acquisition tend to be in the outer layer surrounding the KC and are more abstract. These correlations between the KC and the rest of the GK, and between the GK and the rest of the dictionary as a whole, are binary (0/1), but one can make more graded comparisons by considering definitional chains of increasing lengths. We have accordingly extracted two hierarchies based on degrees of definitional distance, one based on the GK and one based on strongly connected components, to analyze how definitional distance correlates with age of acquisition, concreteness/abstractness and other psycholinguistic variables.

\section{Definitions and Notations}

This section introduces all necessary definitions of graph-theoretical objects studied in this article. The reader is referred to \cite{rosen} for complete graph theory and discrete mathematics introductions. 

\subsection{Graphs}

A \emph{directed graph} is a couple $G = (V,E)$, where $V$ is a finite set of elements called \emph{vertices} and $E \subseteq V \times V$ is a finite set of couples of vertices called \emph{arcs}. Given some graph $G$, we denote its set of vertices and edges by $V(G)$ and $E(G)$ respectively. The \emph{density} of a directed graph is $d(G) = |E(G)| / |V(G)|^2$.

A vertex $u$ is a \emph{predecessor} (or \emph{successor}) of vertex $v$ if $(u,v) \in E$ (or $(v,u) \in E$). The sets of predecessors and successors of $u$ are denoted respectively by $N^-(u)$ and $N^+(v)$. The \emph{in-degree} and \emph{out-degree} of $u$ are defined respectively by $\delta^-(u) = |N^-(u)|$ and $\delta^+(u) = |N^+(u)|$. A vertex of null in-degree (or out-degree) is called a \emph{source} (or \emph{sink}).

A \emph{finite path of length $n$} in a graph is a sequence $(v_0, v_1, \ldots, v_n)$, where $n \geq 0$ is an integer and $(v_{i-1}, v_i) \in E$ for $i=1,2,\ldots,n$. A \emph{$uv$-path} is a path starting with $u$ and ending with $v$. A $uv$-path is a \emph{cycle} if $u = v$. A graph is \emph{acyclic} if it contains no cycles.

Given a graph $G = (V,E)$, we say that $G' = (V',E')$ is a \emph{subgraph of $G$} if $V' \subseteq V$ and $E' \subseteq E$. Moreover, if $V' \subseteq V$, the \emph{subgraph of $G$ induced by $V'$}, denoted by $G[V']$, is the subgraph $G' = (V',E')$ such that $E' = (V' \times V') \cap E$.
\subsection{Dictionaries}

Let $W$ be a finite set whose elements are called \emph{words}, and let
$2^W$ denote the collection of all subsets of $W$. A \emph{dictionary} is a subset $D$ of $W \times 2^W$ such that for every $(w,d_w) \in D$: (i) $d_w \neq \emptyset$ (there is no empty definition) and (ii) $w \notin d_w$ (a word cannot be used to define itself). 

Elements of $D$ are called \emph{entries}. An entry is therefore a couple $(w, d_w)$, where $w$ is a word and $d_w$ is a set of words. The element $w$, the set $d_w$ and the elements of $d_w$ are called respectively the \emph{definiendum} (the defined word), the \emph{definition of $w$} and the \emph{definientes} (the defining words).

\begin{figure}[ht!]
  \centering
  {\small
  \begin{tabular}{m{0.3\linewidth}m{0.63\linewidth}}
  \begin{tabular}{|l|l|}
    \hline
    Word   & Definition   \\
    \hline
    apple  & red fruit    \\
    bad    & not good     \\
    banana & yellow fruit \\
    color  & light dark   \\
    dark   & not light    \\
    edible & good         \\
    fruit  & edible       \\
    good   & not bad      \\
    light  & not dark     \\
    no     & not          \\
    not    & no           \\
    red    & dark color   \\
    tomato & red fruit    \\
    yellow & light color  \\
    \hline
  \end{tabular}
  &
  \begin{tikzpicture}[yscale=0.5, xscale=0.4, pin distance=1.2cm, every pin edge/.style={-, blue}, -stealth, every node/.style={}, every loop/.style={-stealth}, every edge/.style={draw, thick}]
    \node[-, pin={210:\textbf{GK}}] at (0,1) {};
    \fill[fill=blue!10, rotate around={-20:(4,2.3)}] (4,2.3) ellipse (7cm and 3.5cm);
    \node[-, pin={210:\textbf{KC}}] at (3.5,1) {};
    \fill[fill=blue!30, rotate around={90:(4,1.5)}] (4,1.5) ellipse (2.5cm and 1.5cm);

    \node (apple)   at (13, 7)  {apple};
    \node (bad)     at (0, 2)   {bad};
    \node (banana)  at (8, 7.5) {banana};
    \node (color)   at (12, 3)  {color};
    \node (dark)    at (9, 0)   {dark};
    \node (edible)  at (2.5, 9) {edible};
    \node (fruit)   at (8, 9.5) {fruit};
    \node (good)    at (1, 6)   {good};
    \node (light)   at (8, 2.5) {light};
    \node (no)      at (4, 0)   {no};
    \node (not)     at (4, 3)   {not};
    \node (red)     at (16, 4)  {red};
    \node (tomato)  at (15, 9) {tomato};
    \node (yellow)  at (8, 5)   {yellow};
  
    \path (bad)     edge[out=60, in=280]  (good);
    \path (color)   edge                  (red);
    \path (color)   edge                  (yellow);
    \path (dark)    edge                  (color);
    \path (dark)    edge[out=120, in=280] (light);
    \path (dark)    edge                  (red);
    \path (edible)  edge                  (fruit);
    \path (fruit)   edge                  (apple);
    \path (fruit)   edge                  (banana);
    \path (fruit)   edge                  (tomato);
    \path (good)    edge[out=240, in=100] (bad);
    \path (good)    edge                  (edible);
    \path (light)   edge                  (color);
    \path (light)   edge[out=300, in=90]  (dark);
    \path (light)   edge                  (yellow);
    \path (no)      edge[out=60, in=280]  (not);
    \path (not)     edge                  (bad);
    \path (not)     edge                  (dark);
    \path (not)     edge                  (good);
    \path (not)     edge                  (light);
    \path (not)     edge[out=240, in=100] (no);
    \path (red)     edge                  (apple);
    \path (red)     edge                  (tomato);
    \path (yellow)  edge                  (banana);
  \end{tikzpicture}
  \end{tabular}
  }
  \caption{Graph of an artificially contrived (toy) dictionary. First, vertices ``apple", ``banana", ``tomato" are removed, then ``fruit", ``red", ``yellow" and, finally, ``color" and ``edible," leaving the dictionary's Grounding Kernel (GK, paler blue), the subgraph induced by $\{$bad, dark, good, light, no, not$\}$. The strongly-connected Kernel Core (KC, darker blue) is $\{$no, not$\}$. The dictionary has eight distinct minimum grounding sets (MGSs), each containing one and only one word from each of the subsets: $\{$bad, good$\}$,  $\{$dark, light$\}$ and $\{$no, not$\}$. }
  \label{F:graph}
\end{figure}
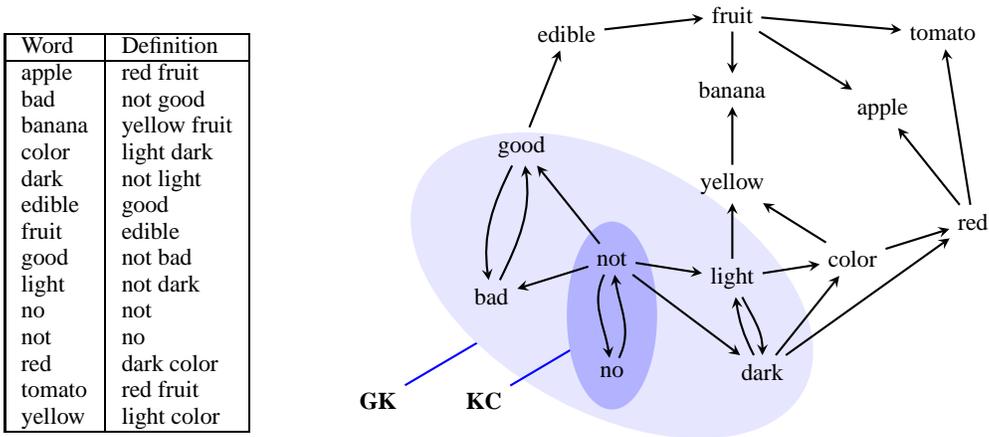

There is a very natural way to derive a graph from a dictionary. The \emph{associated graph} of a dictionary $D \subseteq W \times 2^W$ is the directed graph $G = (V,E)$ where $V = W$ and $(u,v) \in E$ if and only if there exists an entry $(v,d_v) \in D$ such that $u \in d_v$. In fact, associated graphs are exactly directed graphs without loops and without sources. An artificially constructed ``toy" dictionary is illustrated in Figure \ref{F:graph}.

\subsection{Grounding Kernel (GK)}

Let $G = (V,E)$ be a directed graph. We say that $U \subseteq V$ is a \emph{grounding set} (also called \emph{feedback vertex set}) of $G$ if $G[V - U]$ is acyclic, i.e. if $U$ covers every cycle of $G$.

The problem of finding grounding sets of minimum size (MGSs) is NP-complete; hence it is unlikely that one will find an efficient algorithm for solving all instances of this problem. We hope to be able to exploit the particular structure of our graphs to get around this difficulty, and will report on our efforts in a forthcoming paper. Here, we are more interested in extracting hierarchies of definitional distance from dictionary-like graphs.

For this purpose, let $G = (V,E)$ be a directed graph and $\SINKS(G)$ the set of its sinks. We define the operator $\OUTZERO$ by $\OUTZERO(G) = G[V - \SINKS(G)]$. We define $\OUTZERO^n(G)$ as $\OUTZERO^{n-1}(\OUTZERO(G))$ for $n \geq 2$, and it is easily verified that there exists some $\ell$ such that $\OUTZERO^n(G) = \OUTZERO^{\ell}(G)$ for any $n \geq \ell$. Then we
define $\OUTZERO^\infty(G)$ as $\OUTZERO^{\ell}(G)$.

\begin{definition}
The \emph{grounding kernel} of $G$ is given by
$\mbox{GK} = \OUTZERO^\infty(G)$.
\end{definition}

Note that the grounding kernel (GK) of $G$ is well defined for any graph, even if it is acyclic, since the process of removing sources recursively must stop after a finite number of steps. Moreover, the GK is unique for every graph $G$. It is also easy to show that every MGS is included in the GK of $G$. Hence it is of some interest to study the linguistic and cognitive properties of this special set of words as well as its internal graph structure. For instance, we can decompose it according to its strongly connected components (Subsection \ref{SS:core}).

\subsection{Kernel Core (KC)} \label{SS:core}

We recall two classical relations on vertices. Given two vertices $u$ and $v$ of a graph $G$, we write $u \rightarrow v$ if there exists a $uv$-path in $G$ and we write $u \leftrightarrow v$ if $u \rightarrow v$ and $v \rightarrow u$. Note that $\leftrightarrow$ is reflexive, symmetric and transitive so that it is an equivalence relation. Therefore it yields a natural partition of the vertices of $G$. The equivalence classes of this relation are called the \emph{strongly connected components} of $G$.

Let $G$ be a graph and $V_1$, $V_2$, $\ldots$, $V_k$ the strongly connected components of $G$. We construct a graph $G' = (V',E')$ as follows~: $V' = \{V_1,V_2,\ldots,V_k\}$ and $(V_i,V_j) \in E'$ if and only if $V_i \neq V_j$ and there exist $u \in V_i$ and $v \in V_j$ such that $(u,v) \in E$. We call this graph the \emph{SCC-quotient-graph}. The next proposition states a well-known fact about this kind of graph.

\begin{proposition}
Let $G$ be a graph and $G'$ be the SCC-quotient-graph of $G$. Then $G'$
is acyclic. \qed
\end{proposition}
Each acyclic graph induces a partial order on its vertices. In particular, the minimum elements are exactly the sources. They are of particular interest: we define the \emph{kernel core} (KC) as the set of vertices of $G$ belonging to the sources of the SCC-quotient graph $G'$. In the dictionaries we have been studying, there turns out to be only one source.

\section{Hierarchies}

The GK of a graph allows us to divide words into two categories: being in the GK or not. However, we would like to refine this division by introducing hierarchies on the vertices, generating more levels. In this article, we consider two hierarchies. The first is induced by the GK and the second is obtained from the strongly connected components.

Let $G = (V,E)$ be a directed graph associated with some dictionary. Let $K$ be its GK. The GK-level of a vertex $v \in V$ with respect to $K$ is defined by
\begin{equation}
L_{\GK}(v) = \begin{cases}
  0 & \mbox{if $v \in K$,} \\
  \max\{L_{\GK}(u) \mid u \in N^-(v)\} + 1 & \mbox{otherwise.}
\end{cases}
\end{equation}
We will call \emph{GK-hierarchy} the categorization of the vertices of $G$ induced by this level function.

The next hierarchy is based on the strongly connected components. Let $G$ be a graph and $G'$ be its SCC-quotient graph. We define the level $L_{\SCC}(v')$ of a vertex $v'$ of $G'$ as follows
\begin{equation}
L_{\SCC}(v') = \begin{cases}
    0 & \mbox{if $v'$ is a source of $G'$,} \\
    \max\{L_{\SCC}(u') \mid u' \in N^-(v')\} + 1 & \mbox{otherwise.}
\end{cases}
\end{equation}
The level $L_{\SCC}(v)$ of a vertex $v$ of $G$ is the level $L_{\SCC}(v')$ of the strongly connected component $v'$ to which $v$ belongs. This level function is well defined since the graph is acyclic.

The \emph{SCC-hierarchy} is induced by the $L_{\SCC}$ level function on the vertices of $G$. In particular, all elements belonging to the same SCC have the same level.

In the next sections, we study three sets of orderings obtained from those two hierarchies: (1) the ordering induced by the GK-hierarchy on the dictionary as a whole; (2) the ordering induced by the SCC-hierarchy on the dictionary as a whole; (3) the ordering induced by the SCC-hierarchy within the GK alone. We include the third hierarchy for a better understanding of the internal structure of the GK.
\begin{example}
Continuing with the graph of Figure \ref{F:graph}, Table \ref{T:levels} contains the levels of each word according to both hierarchies. Notice that the words ``no" and ``not" have level $0$ while all the other words $u$ satisfy $L_{\SCC}(u) = L_{\GK}(u) + 1$. This is not always the case: this is explained by the fact that the GK of this very small example consists of the KC (corresponding to one connected component) and two other components that are the successors of the GK. In natural language dictionaries, the two level functions may be quite different.
\end{example}
\begin{table}[ht!]
  \centering
  \begin{tabular}{|l|c|c||l|c|c|}
    \hline
    Word   & $L_{\GK}$ & $L_{\SCC}$ & Word   & $L_{\GK}$ & $L_{\SCC}$ \\
    \hline
    apple  & $3$       & $4$        & good   & $0$       & $1$ \\
    bad    & $0$       & $1$        & light  & $0$       & $1$ \\
    banana & $3$       & $4$        & no     & $0$       & $0$ \\
    color  & $1$       & $2$        & not    & $0$       & $0$ \\
    dark   & $0$       & $1$        & red    & $2$       & $3$ \\
    edible & $1$       & $2$        & tomato & $3$       & $4$ \\
    fruit  & $2$       & $3$        & yellow & $2$       & $3$ \\
    \hline
  \end{tabular}
  \caption{Levels corresponding respectively to the GK-hierarchy and to the SCC-hierarchy.}
  \label{T:levels}
\end{table}

\section{Natural Language Dictionaries}

We have applied the above hierarchies to the study of two dictionaries: CIDE \cite{cide} and LDOCE \cite{ldoce}.  As in many natural language processing problems, we are confronted with variation in words' morphosyntactic form and with polysemy (multiple meanings for the same word-form). Morphosyntactic variation was removed using Porter's algorithm \cite{porter}. To reduce polysemy we applied a common approximation: we kept only the first definition for each word. Finally, we removed loops and words with empty definitions. The number of vertices, the number of edges and the density of the two dictionaries, along with those of their GK and their ``kernel core" (KC, the subgraph induced by their larger strongly connected component, defined below), are represented in Table \ref{T:features}.

\begin{table}[ht!]
  \centering
  \begin{tabular}{|c|c|c|c|c|c|c|}
    \hline
    Dictionary  & $\CIDE$    & $\LDOCE$   & $\GK_{\CIDE}$ & $\GK_{\LDOCE}$ & $\KC_{\CIDE}$ & $\KC_{\LDOCE}$ \\
    \hline
    Number of vertices    & $19 053$   & $23 998$   & $1725$        & $2001$         & $1453$        & $1371$ \\
    Number of edges & $221 374$  & $239 149$  & $20718$       & $20794$        & $16871$       & $14 062$ \\
    Density     & $0.000610$ & $0.000415$ & $0.00696$     & $0.00519$      & $0.00799$     & $0.00748$ \\
    \hline
  \end{tabular}
  \caption{Main features of the complete dictionaries CIDE and LDOCE, their grounding kernels (GKs) and their kernel cores (KCs).}
  \label{T:features}
\end{table}

Like many networks derived from natural models, dictionary graphs satisfy almost all criteria of small-world graphs \cite{tenenbaum}. The two graphs are sparse and have density lower than $1\%$. Both dictionaries turn out to contain one single strongly connected subset. Moreover, their in-degree and out-degree distribution seem to follow a normal law and power-law, respectively. However, they also have a large number of small strongly connected components. The size of CIDE's biggest SCC is $1453$ and LDOCE's is $1371$ (see Table \ref{T:features}), while the remaining SCCs are very small. 

A remarkable observation is that the KC of both dictionaries is obtained from only a single source, and that source corresponds to the biggest strongly connected component. Hence, the heart of the cyclic structure of CIDE and LDOCE is found in the KC.

\section{Psycholinguistic Correlates of Words in the GK, KC, and Definitional Hierarchies}

We analyzed how the structure of dictionary definition space -- in terms of GK, the KC, and the two hierarchies of definitional distance that they induce -- is related to five psycholinguistic variables: 

\textbf{AOA: Age of Acquisition} - the age at which a word is learned\\
\textbf{C: Concreteness} - degree of concreteness/abstractness of word's referent\\
\textbf{I: Imageability} - how readily one could generate a mental (visual) image of word's referent\\
\textbf{BF: Brown Frequency} - spoken frequency of word\\
\textbf{TLF: Thorndike-Lodge Frequency} - written frequency of word

The psycholinguistic variables (AOA, C, and I) came from the MRC psycholinguistic database \cite{mrc}. Because MRC only covered 10\% of the words in CIDE (and 8\% of LDOCE), we merged each with further databases that were highly correlated with the MRC psycholinguistic database and to increase coverage to 33\% for CIDE and 26\% for LDOCE. \cite{cortese} \cite{stadthagen}

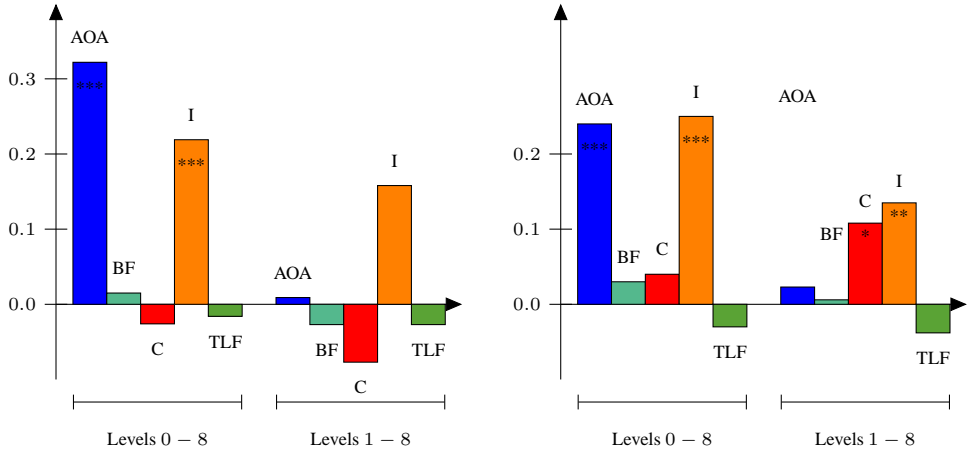
\begin{figure}[ht!]
    {\scriptsize
    \centering
    \begin{tabular}{m{.45\linewidth}m{.45\linewidth}}
    \input{gk_hierarchy_multiple_regression}
    & \input{scc_hierarchy_multiple_regression}
    \end{tabular}
    }
    \caption{Beta values from multiple regression of the 5 psycholinguistic variables (AOA, C, I, BF, TLF) against the multiple levels of the hierarchies in definition space induced by GK (left) and SCCs (right) (for CIDE; LDOCE pattern was the same). \textbf{Left, GK Hierarchy:} For the GK hierarchy, significant effects occur only in the transition from the level 0 -- GK itself -- to level 1 (Levels 0-8, left): GK words are acquired younger and are more imageable; but if the 0 level (GK) is excluded from the analysis (Levels 1-8, right), there is no longer any significant correlation. \textbf{Right, SCC Hierarchy:}  For the SSC hierarchy, KC words are acquired younger, less concrete and less imageable; again, the locus of the AOA (age) correlation is just in the transition from level 0 (KC) to level 1, whereas the C and I correlation continues through all the levels. To visualize the levels, see Figures \ref{F:sccdictio} and \ref{F:sccgk}.)}
    \label{F:regressiongk}
\end{figure}


We did statistical analyses of the three hierarchies that were induced (1) by the GK  across the entire dictionary, (2) by the SCCs (Strongly Connected Components) across the entire dictionary and (3) by the SCCs within the GK only.

\begin{figure}[hb!]
  \centering
  {\scriptsize
  \begin{tabular}{cc}
  \input{scc_hierarchy_whole_cide}
  &
  \input{scc_hierarchy_whole_ldoce}
  \end{tabular}
  }
  \caption{Means for AOA, C, I, BF, and TLF for each level of CIDE (left) and LDOCE (right) with respect to the \emph{SCC-induced hierarchical levels across the entire dictionary}. In the SCC hierarchy, concreteness, imageability, and both oral and written word frequency decline, whereas after its initial increase from the KC to level 1, age of acquisition stays flat. See text for ANOVA and Factor Analysis}
  \label{F:sccdictio}
\end{figure}
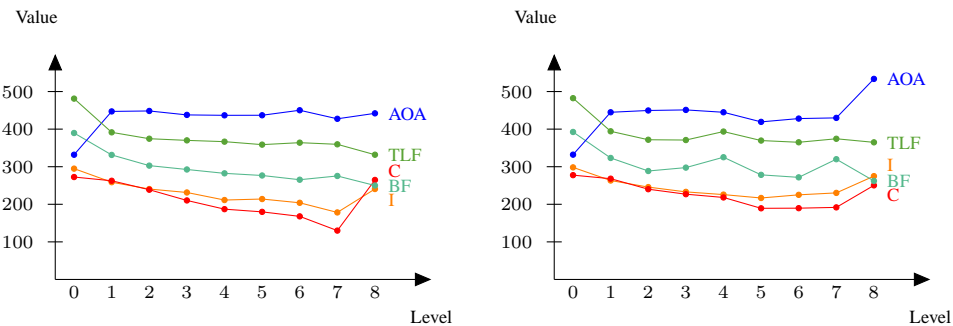

With the GK-induced hierarchy, we did two linear regression analyses. Level $0$ words (i.e., those within the GK) were included in the first analysis and excluded from the second. Figure \ref{F:regressiongk} shows that in the first analysis AOA and I are significantly correlated with definitional distance. In the second, no correlation is significant. Hence the only significant effect here is the difference between the GK and the rest of the dictionary: GK words are learned earlier and are more imageable; this effect does not carry over to the higher levels in the induced hierarchy.

In contrast, in the analyses for the SCC-induced hierarchy across the entire dictionary, an ANOVA showed that differences in C and in I at higher levels of the hierarchy are significant too; in post-hoc tests, levels $0$, $2$, $3$ and $4$ differed from level $1$. The means for each level and each variable are shown in Figure \ref{F:sccdictio}.

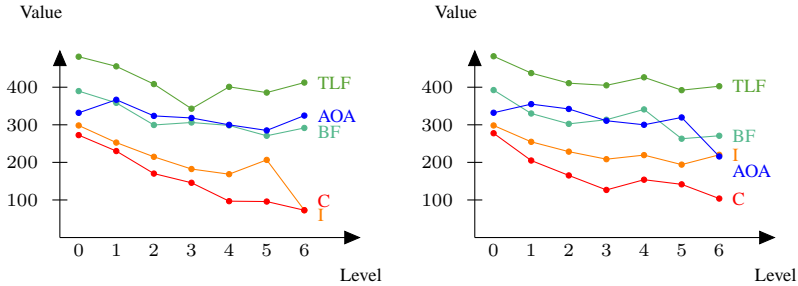
\begin{figure}[ht!]
  \centering
  {\scriptsize
  \input{scc_hierarchy_gk_cide}
  \quad
  \input{scc_hierarchy_gk_ldoce}
  }
  \caption{Means for AOA, C, I, BF and TLF for each level of CIDE (left) and LDOCE (right) with respect to the \emph{SCC-induced hierarchical levels within the GK.} If we induce the hierarchy within the GK alone, the pattern is similar to the external hierarchy: C, I and both frequencies decline with increasing definitional distance from the KC, while AOA (age), after the initial transition from KC to level 1 again remains flat.}
  \label{F:sccgk}
\end{figure}

Finally, for the SCC-induced hierarchy within GK alone, AOA and I are significant in the multiple regression (see Figure \ref{F:sccgk}). Moreover, all variables had significant effects in the ANOVA, with the more specific post-hoc tests showing that C and I change from level to level, and AOA differs for most levels.

Age of acquisition is earlier for level $0$ words (the KC) compared to the other levels in the hierarchy induced by the SCCs for the dictionary as a whole. Words in the KC are also more abstract and less imageable than those in the other SCCs. KC words are also used more frequently, both orally and in writing. These results are similar to those reported by \cite{tenenbaum}.

Within the GK alone, words become less concrete, imageable and frequent along the SCC-induced levels starting from KC. Age of acquisition is also significantly older than KC, but only for the first level, after which age remains flat. The correlation between concreteness and age of acquisition is also significantly higher for GK words outside than inside the KC.

These results cast further light on the prior finding of \cite{chicoisne} that GK words are learned earlier and more concrete, but that when AOA's covariance with C is partialled out, C's polarity shifts:  GK words that are not learned earlier are significantly more abstract than the rest of the dictionary. Our newer finding that the correlation between AOA and C is higher within the GK layer outside the KC suggests that the outer layer may be the locus of this difference. The GK's KC is more concrete and learned earlier, whereas its outer layer is more abstract, and unrelated to age of acquisition.

In the GK-induced hierarchy, only age and imageability were significant. This seemed to contradict our previous findings \cite{chicoisne} comparing the GK with the rest of the dictionary, so we reanalyzed this hierarchy excluding its bottom level, the GK. This eliminated all correlations (Figure \ref{F:regressiongk}).

\section{Concluding Remarks}

The GK is a subset of the dictionary with features that differ substantially from the rest of the dictionary. The factors underlying  the polarity change in concreteness observed in previous work \cite{chicoisne}  -- with the GK being more concrete and learned earlier than the rest of the dictionary, but more abstract when the covariance with age is partialled out  -- has now been further refined: It turns out that the GK consists of a large, strongly connected KC plus a smaller, less interconnected outer layer. The KC, like the GK, is more concrete and learned earlier than the rest of the dictionary, and it is also more concrete and learned earlier than the outer layer. But with the KC, when the effects of age of acquisition are partialled out, there is no polarity reversal: The KC remains more concrete than its outer layer and also remains more concrete than the rest of the dictionary. So it is the outer layer that is more abstract than the KC , and hence the polarity reversal is related to the difference between the KC and the outer layer of the GK.

\begin{figure}[ht!]
  \centering
  \input{general_schematic}
  \caption{Summary of findings: Words in the Kernel Core (KC) are more concrete, imageable, and frequent, and learned younger. The effect is graded by definitional distance for concreteness and imageability but dichotomous for age and word frequency.}
  \label{F:schematic}
\end{figure}
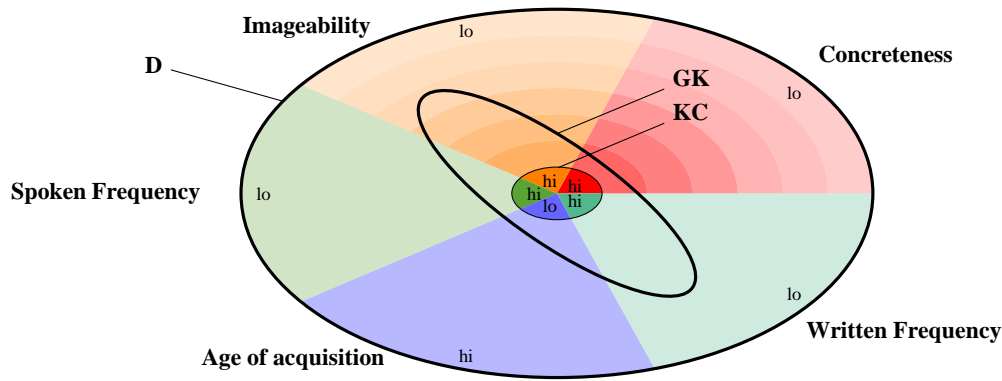

To further refine the differences between the GK's KC, the GK's outer layer and the rest of the dictionary, we induced three orderings in definitional space to produce a graded series of hierarchical levels at an increasing definitional distance from its bottom level or source, to test whether the dichotomous effects based on the GK or the KC extended beyond, along a graded series of definitional distances. One hierarchy was induced from the GK; the second was induced from the entire dictionary's strongly connected components (SCCs) and the third was induced from the SCCs within the GK alone. The bottom level or ``source" of the GK-based hierarchy of definitional distances was the GK itself; the bottom level of both the SCC hierarchies turned out to be the GK's large strongly connected KC. The successive levels of the SCC-induced hierarchy beyond the KC but within the GK and the levels on beyond the GK into the definitional space of the rest of the dictionary both turned out to be significantly correlated with the psycholinguistic variables (age, concreteness, imageability, oral and written frequency): Concreteness and imageability continue to decrease all the way out to the periphery in definition space; oral and written word frequency likewise continue to decrease; but the correlation with age of acquisition is present only in the contrast between the KC and the next level, which is the outer layer of the GK. The effects of SCC-induction for the entire dictionary were also similar to those of \cite{tenenbaum}: In their small-world analyses too, the KC was acquired earlier then the rest of the corpus and more frequently used (see Summary Figure \ref{F:schematic}).

All categories, even the most ``concrete" are in fact abstractions, because we must abstract from particular cases, even concrete sensorimotor ones, in order to find the invariant features that distinguish category members from nonmembers and allow us to do the right thing with the right kind of thing. But the more that categories are based on other categories, the more abstract they become, and this is reflected by the distances in our induced definitional space. It is in the nature of words to be amenable to combination and recombination in such a way as to define or describe ever more categories. Defining, like eating, is something we \emph{do}. Our more concrete categories are answerable to the constraints of the sensorimotor world in which they are grounded, but our more abstract categories are increasingly answerable only to combinations of other categories, as we describe and define them. In abstract mathematics, that constraint, though only formal, is still a rigorous one. In more hermeneutic discourse (e.g., constitutional law or theology) the main constraint on words increasingly becomes just other words. Our mental lexicon must encode the meaning of all the words we use in our thought and discourse. Hierarchies in dictionary space may turn out to have counterparts in cognitive space.

\end{document}

%% file: gk_hierarchy_multiple_regression.tex
\begin{tikzpicture}[-triangle 45, yscale=10, xscale=0.45]
\draw (0,-0.1) -- (0,0.4);
\draw (0,0) -- (12.0000000000000,0);
\draw[-] (-0.333333,-0.0) -- (0.333333,-0.0);
\node at (-1.0,-0.0) {$0.0$};
\draw[-] (-0.333333,0.1) -- (0.333333,0.1);
\node at (-1.0,0.1) {$0.1$};
\draw[-] (-0.333333,0.2) -- (0.333333,0.2);
\node at (-1.0,0.2) {$0.2$};
\draw[-] (-0.333333,0.3) -- (0.333333,0.3);
\node at (-1.0,0.3) {$0.3$};
\draw[fillaoa] (0.500000000000000,0) rectangle (1.50000000000000,0.322000000000000);
\draw[fillbf] (1.50000000000000,0) rectangle (2.50000000000000,0.0150000000000000);
\draw[fillc] (2.50000000000000,0) rectangle (3.50000000000000,-0.0260000000000000);
\draw[filli] (3.50000000000000,0) rectangle (4.50000000000000,0.219000000000000);
\draw[filltlf] (4.50000000000000,0) rectangle (5.50000000000000,-0.0160000000000000);
\draw[fillaoa] (6.50000000000000,0) rectangle (7.50000000000000,0.00900000000000000);
\draw[fillbf] (7.50000000000000,0) rectangle (8.50000000000000,-0.0270000000000000);
\draw[fillc] (8.50000000000000,0) rectangle (9.50000000000000,-0.0770000000000000);
\draw[filli] (9.50000000000000,0) rectangle (10.5000000000000,0.158000000000000);
\draw[filltlf] (10.5000000000000,0) rectangle (11.5000000000000,-0.0270000000000000);
\node at (1.00000000000000,0.355333000000000) {AOA};
\node at (2.00000000000000,0.0483330000000000) {BF};
\node at (3.00000000000000,-0.0593330000000000) {C};
\node at (4.00000000000000,0.252333000000000) {I};
\node at (5.00000000000000,-0.0493330000000000) {TLF};
\node at (7.00000000000000,0.0423330000000000) {AOA};
\node at (8.00000000000000,-0.0603330000000000) {BF};
\node at (9.00000000000000,-0.110333000000000) {C};
\node at (10.0000000000000,0.191333000000000) {I};
\node at (11.0000000000000,-0.0603330000000000) {TLF};
\node at (1.00000000000000,0.288667) {***};
\node at (4.00000000000000,0.185667) {***};
\node at (3.0, -0.18) {Levels $0-8$};
\node at (9.0, -0.18) {Levels $1-8$};
\draw[|-|] (0.5, -0.13) -- (5.5, -0.13);
\draw[|-|] (6.5, -0.13) -- (11.5, -0.13);
\end{tikzpicture}

%% file: scc_hierarchy_multiple_regression.tex
\begin{tikzpicture}[-triangle 45, yscale=10, xscale=0.45]
\draw (0,-0.1) -- (0,0.4);
\draw (0,0) -- (12.0000000000000,0);
\draw[-] (-0.333333,-0.0) -- (0.333333,-0.0);
\node at (-1.0,-0.0) {$0.0$};
\draw[-] (-0.333333,0.1) -- (0.333333,0.1);
\node at (-1.0,0.1) {$0.1$};
\draw[-] (-0.333333,0.2) -- (0.333333,0.2);
\node at (-1.0,0.2) {$0.2$};
\draw[fillaoa] (0.500000000000000,0) rectangle (1.50000000000000,0.240000000000000);
\draw[fillbf] (1.50000000000000,0) rectangle (2.50000000000000,0.0300000000000000);
\draw[fillc] (2.50000000000000,0) rectangle (3.50000000000000,0.0400000000000000);
\draw[filli] (3.50000000000000,0) rectangle (4.50000000000000,0.250000000000000);
\draw[filltlf] (4.50000000000000,0) rectangle (5.50000000000000,-0.0300000000000000);
\draw[fillaoa] (6.50000000000000,0) rectangle (7.50000000000000,0.023000000000000);
\draw[fillbf] (7.50000000000000,0) rectangle (8.50000000000000,0.0060000000000000);
\draw[fillc] (8.50000000000000,0) rectangle (9.50000000000000,0.108000000000000);
\draw[filli] (9.50000000000000,0) rectangle (10.5000000000000,0.135000000000000);
\draw[filltlf] (10.5000000000000,0) rectangle (11.5000000000000,-0.0380000000000000);
\node at (1.00000000000000,0.273333000000000) {AOA};
\node at (2.00000000000000,0.0633330000000000) {BF};
\node at (3.00000000000000,0.0733330000000000) {C};
\node at (4.00000000000000,0.283333000000000) {I};
\node at (5.00000000000000,-0.0633330000000000) {TLF};
\node at (7.00000000000000,0.277333000000000) {AOA};
\node at (8.00000000000000,0.0933330000000000) {BF};
\node at (9.00000000000000,0.137333000000000) {C};
\node at (10.0000000000000,0.164333000000000) {I};
\node at (11.0000000000000,-0.0713330000000000) {TLF};
\node at (1.00000000000000,0.206667) {***};
\node at (4.00000000000000,0.216667) {***};
\node at (9.00000000000000,0.090667) {*};
\node at (10.0000000000000,0.117667) {**};
\node at (3.0, -0.18) {Levels $0-8$};
\node at (9.0, -0.18) {Levels $1-8$};
\draw[|-|] (0.5, -0.13) -- (5.5, -0.13);
\draw[|-|] (6.5, -0.13) -- (11.5, -0.13);
\end{tikzpicture}

%% file: scc_hierarchy_whole_cide.tex
\begin{tikzpicture}[scale=0.5, -triangle 45]
\draw (0,0.0) -- (0,6.0);
\draw (0,0) -- (10.0000000000000,0);
\node at (0.500000000000000,-0.333333) {$0$};
\node at (1.50000000000000,-0.333333) {$1$};
\node at (2.50000000000000,-0.333333) {$2$};
\node at (3.50000000000000,-0.333333) {$3$};
\node at (4.50000000000000,-0.333333) {$4$};
\node at (5.50000000000000,-0.333333) {$5$};
\node at (6.50000000000000,-0.333333) {$6$};
\node at (7.50000000000000,-0.333333) {$7$};
\node at (8.50000000000000,-0.333333) {$8$};
\draw[-] (-0.2,1.0) -- (0.2,1.0);
\node at (-1.0,1.0) {$100$};
\draw[-] (-0.2,2.0) -- (0.2,2.0);
\node at (-1.0,2.0) {$200$};
\draw[-] (-0.2,3.0) -- (0.2,3.0);
\node at (-1.0,3.0) {$300$};
\draw[-] (-0.2,4.0) -- (0.2,4.0);
\node at (-1.0,4.0) {$400$};
\draw[-] (-0.2,5.0) -- (0.2,5.0);
\node at (-1.0,5.0) {$500$};
\node at (10.0000000000000,-1.0) {Level};
\node at (-0.500000000000000,7.0) {Value};
\draw[-, i] plot[mark=*] coordinates { (0.500000000000000,2.9489) (1.50000000000000,2.5877) (2.50000000000000,2.4058) (3.50000000000000,2.3144) (4.50000000000000,2.1132) (5.50000000000000,2.1399) (6.50000000000000,2.0399) (7.50000000000000,1.7834) (8.50000000000000,2.4069) };
\draw[-, c] plot[mark=*] coordinates { (0.500000000000000,2.7254) (1.50000000000000,2.6259) (2.50000000000000,2.3877) (3.50000000000000,2.1026) (4.50000000000000,1.8704) (5.50000000000000,1.7984) (6.50000000000000,1.6787) (7.50000000000000,1.3) (8.50000000000000,2.6475) };
\draw[-, bf] plot[mark=*] coordinates { (0.500000000000000,3.89707251441775) (1.50000000000000,3.31341339735847) (2.50000000000000,3.02890510010872) (3.50000000000000,2.92620873635117) (4.50000000000000,2.82352287440518) (5.50000000000000,2.76636737720574) (6.50000000000000,2.65384708792860) (7.50000000000000,2.75376048027328) (8.50000000000000,2.50000000000000) };
\draw[-, aoa] plot[mark=*] coordinates { (0.500000000000000,3.3185) (1.50000000000000,4.4712) (2.50000000000000,4.4835) (3.50000000000000,4.3785) (4.50000000000000,4.3673) (5.50000000000000,4.3689) (6.50000000000000,4.5022) (7.50000000000000,4.275) (8.50000000000000,4.42) };
\draw[-, tlf] plot[mark=*] coordinates { (0.500000000000000,4.81194338562722) (1.50000000000000,3.91442254672521) (2.50000000000000,3.74493378073306) (3.50000000000000,3.70099754915048) (4.50000000000000,3.66656489477275) (5.50000000000000,3.58746223724141) (6.50000000000000,3.63972347392360) (7.50000000000000,3.59582020224327) (8.50000000000000,3.31894350786512) };
\node[label={[aoa]right:AOA}] at (8.50000000000000,4.42) {};
\node[label={[bf]right:BF}] at (8.50000000000000,2.50000000000000) {};
\node[label={[c]right:C}] at (8.50000000000000,2.8875) {};
\node[label={[i]right:I}] at (8.50000000000000,2.1069) {};
\node[label={[tlf]right:TLF}] at (8.50000000000000,3.31894350786512) {};
\end{tikzpicture}

%% file: scc_hierarchy_whole_ldoce.tex
\begin{tikzpicture}[scale=0.5, -triangle 45]
\draw (0,0.0) -- (0,6.0);
\draw (0,0) -- (10.0000000000000,0);
\node at (0.500000000000000,-0.333333) {$0$};
\node at (1.50000000000000,-0.333333) {$1$};
\node at (2.50000000000000,-0.333333) {$2$};
\node at (3.50000000000000,-0.333333) {$3$};
\node at (4.50000000000000,-0.333333) {$4$};
\node at (5.50000000000000,-0.333333) {$5$};
\node at (6.50000000000000,-0.333333) {$6$};
\node at (7.50000000000000,-0.333333) {$7$};
\node at (8.50000000000000,-0.333333) {$8$};
\draw[-] (-0.2,1.0) -- (0.2,1.0);
\node at (-1.0,1.0) {$100$};
\draw[-] (-0.2,2.0) -- (0.2,2.0);
\node at (-1.0,2.0) {$200$};
\draw[-] (-0.2,3.0) -- (0.2,3.0);
\node at (-1.0,3.0) {$300$};
\draw[-] (-0.2,4.0) -- (0.2,4.0);
\node at (-1.0,4.0) {$400$};
\draw[-] (-0.2,5.0) -- (0.2,5.0);
\node at (-1.0,5.0) {$500$};
\node at (10.0000000000000,-1.0) {Level};
\node at (-0.500000000000000,7.0) {Value};
\draw[-, i] plot[mark=*] coordinates { (0.500000000000000,2.9824) (1.50000000000000,2.6329) (2.50000000000000,2.4575) (3.50000000000000,2.3303) (4.50000000000000,2.2577) (5.50000000000000,2.1678) (6.50000000000000,2.2516) (7.50000000000000,2.3032) (8.50000000000000,2.7495) };
\draw[-, c] plot[mark=*] coordinates { (0.500000000000000,2.7754) (1.50000000000000,2.6813) (2.50000000000000,2.4015) (3.50000000000000,2.269) (4.50000000000000,2.1819) (5.50000000000000,1.8932) (6.50000000000000,1.8967) (7.50000000000000,1.9177) (8.50000000000000,2.5033) };
\draw[-, bf] plot[mark=*] coordinates { (0.500000000000000,3.92436528645618) (1.50000000000000,3.23270411061076) (2.50000000000000,2.88594220775130) (3.50000000000000,2.97615769104603) (4.50000000000000,3.25239157681570) (5.50000000000000,2.78317176967214) (6.50000000000000,2.71681179484045) (7.50000000000000,3.19964316857068) (8.50000000000000,2.62163953243245) };
\draw[-, aoa] plot[mark=*] coordinates { (0.500000000000000,3.3206) (1.50000000000000,4.4486) (2.50000000000000,4.4961) (3.50000000000000,4.5114) (4.50000000000000,4.4486) (5.50000000000000,4.1929) (6.50000000000000,4.2809) (7.50000000000000,4.2993) (8.50000000000000,5.3367) };
\draw[-, tlf] plot[mark=*] coordinates { (0.500000000000000,4.82376436164599) (1.50000000000000,3.94235983060996) (2.50000000000000,3.71745973467572) (3.50000000000000,3.70968755134573) (4.50000000000000,3.93539631638857) (5.50000000000000,3.69602780546265) (6.50000000000000,3.64995886746465) (7.50000000000000,3.74341623086415) (8.50000000000000,3.64859241894673) };
\node[label={[aoa]right:AOA}] at (8.50000000000000,5.3367) {};
\node[label={[bf]right:BF}] at (8.50000000000000,2.62163953243245) {};
\node[label={[c]right:C}] at (8.50000000000000,2.2633) {};
\node[label={[i]right:I}] at (8.50000000000000,3.0495) {};
\node[label={[tlf]right:TLF}] at (8.50000000000000,3.64859241894673) {};
\end{tikzpicture}

%% file: scc_hierarchy_gk_cide.tex
\begin{tikzpicture}[scale=0.5, -triangle 45]
\draw (0,0.0) -- (0,5.0);
\draw (0,0) -- (8.00000000000000,0);
\node at (0.500000000000000,-0.333333) {$0$};
\node at (1.50000000000000,-0.333333) {$1$};
\node at (2.50000000000000,-0.333333) {$2$};
\node at (3.50000000000000,-0.333333) {$3$};
\node at (4.50000000000000,-0.333333) {$4$};
\node at (5.50000000000000,-0.333333) {$5$};
\node at (6.50000000000000,-0.333333) {$6$};
\draw[-] (-0.2,1.0) -- (0.2,1.0);
\node at (-1.0,1.0) {$100$};
\draw[-] (-0.2,2.0) -- (0.2,2.0);
\node at (-1.0,2.0) {$200$};
\draw[-] (-0.2,3.0) -- (0.2,3.0);
\node at (-1.0,3.0) {$300$};
\draw[-] (-0.2,4.0) -- (0.2,4.0);
\node at (-1.0,4.0) {$400$};
\node at (8.00000000000000,-1.0) {Level};
\node at (-0.500000000000000,6.0) {Value};
\draw[-, i] plot[mark=*] coordinates { (0.500000000000000,2.9824) (1.50000000000000,2.5271) (2.50000000000000,2.148) (3.50000000000000,1.8231) (4.50000000000000,1.6857) (5.50000000000000,2.0661) (6.50000000000000,0.7275) };
\draw[-, c] plot[mark=*] coordinates { (0.500000000000000,2.7254) (1.50000000000000,2.3013) (2.50000000000000,1.7019) (3.50000000000000,1.4574) (4.50000000000000,0.9678) (5.50000000000000,0.9579) (6.50000000000000,0.7262) };
\draw[-, bf] plot[mark=*] coordinates { (0.500000000000000,3.89707251441775) (1.50000000000000,3.58351851847766) (2.50000000000000,2.99517395672958) (3.50000000000000,3.06154065307048) (4.50000000000000,2.98283137373023) (5.50000000000000,2.70794415416798) (6.50000000000000,2.91588830833597) };
\draw[-, aoa] plot[mark=*] coordinates { (0.500000000000000,3.3185) (1.50000000000000,3.6665) (2.50000000000000,3.2371) (3.50000000000000,3.182) (4.50000000000000,2.999) (5.50000000000000,2.85) (6.50000000000000,3.245) };
\draw[-, tlf] plot[mark=*] coordinates { (0.500000000000000,4.81194338562722) (1.50000000000000,4.55512046558565) (2.50000000000000,4.08331192379317) (3.50000000000000,3.42840166642267) (4.50000000000000,4.00973860459671) (5.50000000000000,3.85751324245264) (6.50000000000000,4.12349381555651) };
\node[label={[aoa]right:AOA}] at (6.50000000000000,3.245) {};
\node[label={[bf]right:BF}] at (6.50000000000000,2.81588830833597) {};
\node[label={[c]right:C}] at (6.50000000000000,0.9762) {};
\node[label={[i]right:I}] at (6.50000000000000,0.5975) {};
\node[label={[tlf]right:TLF}] at (6.50000000000000,4.12349381555651) {};
\end{tikzpicture}

%% file: scc_hierarchy_gk_ldoce.tex
\begin{tikzpicture}[scale=0.5, -triangle 45]
\draw (0,0.0) -- (0,5.0);
\draw (0,0) -- (8.00000000000000,0);
\node at (0.500000000000000,-0.333333) {$0$};
\node at (1.50000000000000,-0.333333) {$1$};
\node at (2.50000000000000,-0.333333) {$2$};
\node at (3.50000000000000,-0.333333) {$3$};
\node at (4.50000000000000,-0.333333) {$4$};
\node at (5.50000000000000,-0.333333) {$5$};
\node at (6.50000000000000,-0.333333) {$6$};
\draw[-] (-0.2,1.0) -- (0.2,1.0);
\node at (-1.0,1.0) {$100$};
\draw[-] (-0.2,2.0) -- (0.2,2.0);
\node at (-1.0,2.0) {$200$};
\draw[-] (-0.2,3.0) -- (0.2,3.0);
\node at (-1.0,3.0) {$300$};
\draw[-] (-0.2,4.0) -- (0.2,4.0);
\node at (-1.0,4.0) {$400$};
\node at (8.00000000000000,-1.0) {Level};
\node at (-0.500000000000000,6.0) {Value};
\draw[-, i] plot[mark=*] coordinates { (0.500000000000000,2.9824) (1.50000000000000,2.5471) (2.50000000000000,2.2857) (3.50000000000000,2.0851) (4.50000000000000,2.1936) (5.50000000000000,1.9406) (6.50000000000000,2.2011) };
\draw[-, c] plot[mark=*] coordinates { (0.500000000000000,2.7754) (1.50000000000000,2.0497) (2.50000000000000,1.6527) (3.50000000000000,1.267) (4.50000000000000,1.5389) (5.50000000000000,1.4156) (6.50000000000000,1.0375) };
\draw[-, bf] plot[mark=*] coordinates { (0.500000000000000,3.92436528645618) (1.50000000000000,3.30058483955704) (2.50000000000000,3.02475995644278) (3.50000000000000,3.13487665444077) (4.50000000000000,3.40947457793431) (5.50000000000000,2.62953472492766) (6.50000000000000,2.70794415416798) };
\draw[-, aoa] plot[mark=*] coordinates { (0.500000000000000,3.3206) (1.50000000000000,3.5504) (2.50000000000000,3.4216) (3.50000000000000,3.1077) (4.50000000000000,3.0004) (5.50000000000000,3.1956) (6.50000000000000,2.1575) };
\draw[-, tlf] plot[mark=*] coordinates { (0.500000000000000,4.82376436164599) (1.50000000000000,4.37624093696668) (2.50000000000000,4.10762612060556) (3.50000000000000,4.05010361973101) (4.50000000000000,4.26345515819470) (5.50000000000000,3.92156922074759) (6.50000000000000,4.02627890056972) };
\node[label={[aoa]right:AOA}] at (6.50000000000000,1.7675) {};
\node[label={[bf]right:BF}] at (6.50000000000000,2.70794415416798) {};
\node[label={[c]right:C}] at (6.50000000000000,1.0375) {};
\node[label={[i]right:I}] at (6.50000000000000,2.2011) {};
\node[label={[tlf]right:TLF}] at (6.50000000000000,4.02627890056972) {};
\end{tikzpicture}

%% file: general_schematic.tex
\begin{tikzpicture}[xscale=0.3, yscale=0.35, zone/.style={rounded corners, fill=black}, pin distance=15mm, every pin edge/.style={black}]
  \foreach \a/\o in {0/1,1/0.6,2/0.5,3/0.4,4/0.3,5/0.25,6/0.2} {
    \fill[c, opacity=\o] (0:2*\a cm and \a cm) arc (0:72:2*\a cm and \a cm) -- (72:2*\a cm+2 cm and \a cm+1 cm) arc (72:0:2*\a cm+2 cm and \a cm+1 cm);
    \fill[i, opacity=\o] (72:2*\a cm and \a cm) arc (72:144:2*\a cm and \a cm) -- (144:2*\a cm+2 cm and \a cm+1 cm) arc (144:72:2*\a cm+2 cm and \a cm+1 cm);
  }
  \fill[tlf, opacity=1] (144:0cm and 0cm) arc (144:216:0cm and 0cm) -- (216:2cm and 1cm) arc (216:144:2cm and 1cm);
  \fill[tlf, opacity=0.28] (144:2cm and 1cm) arc (144:216:2cm and 1cm) -- (216:14cm and 7cm) arc (216:144:14cm and 7cm);
  \fill[aoa, opacity=0.6] (216:0cm and 0cm) arc (216:288:0cm and 0cm) -- (288:2cm and 1cm) arc (288:216:2cm and 1cm);
  \fill[aoa, opacity=0.28] (216:2cm and 1cm) arc (216:288:2cm and 1cm) -- (288:14cm and 7cm) arc (288:216:14cm and 7cm);
  \fill[bf, opacity=1] (288:0cm and 0cm) arc (288:360:0cm and 0cm) -- (360:2cm and 1cm) arc (360:288:2cm and 1cm);
  \fill[bf, opacity=0.28] (288:2cm and 1cm) arc (288:360:2cm and 1cm) -- (360:14cm and 7cm) arc (360:288:14cm and 7cm);

  {\scriptsize
  \node at (36:1cm and 0.5cm) {hi};
  \node at (108:1cm and 0.5cm) {hi};
  \node at (180:1cm and 0.5cm) {hi};
  \node at (252:1cm and 0.5cm) {lo};
  \node at (324:1cm and 0.5cm) {hi};
  \node at (36:13cm and 6.5cm) {lo};
  \node at (108:13cm and 6.5cm) {lo};
  \node at (180:13cm and 6.5cm) {lo};
  \node at (252:13cm and 6.5cm) {hi};
  \node at (324:13cm and 6.5cm) {lo};
  }
  
  {\footnotesize
  \node at (36:18cm and 9cm) {\textbf{Concreteness}};
  \node at (128:18cm and 8cm) {\textbf{Imageability}};
  \node at (180:20cm and 9cm) {\textbf{Spoken Frequency}};
  \node at (232:19cm and 8cm) {\textbf{Age of acquisition}};
  \node at (324:19cm and 9cm) {\textbf{Written Frequency}};

  \draw[very thick, rotate=-30] (0,0) ellipse (7cm and 2cm);
  \draw[very thick] (0,0) ellipse (14cm and 7cm);
  \draw (0,0) ellipse (2cm and 1cm);
  \node[outer sep=0pt, inner sep=0pt, pin={20:\textbf{GK}}] at (0,2.27) {};
  \node[outer sep=0pt, inner sep=0pt, pin distance=5cm, pin={20:\textbf{KC}}] at (0,0.96) {};
  \node[outer sep=0pt, inner sep=0pt, pin distance=5cm, pin={170:\textbf{D}}] at (150:14cm and 7cm) {};
  }
\end{tikzpicture}

%% file: nips09.bbl
\begin{thebibliography}{2}
\bibitem{textgraphs} Blondin Mass\'e, A., Chicoisne, G., Gargouri, Y., Harnad, S., Marcotte, O., Picard, O. \emph{How is meaning grounded in dictionary definitions?}, In TextGraphs '08: Proceedings of the 3rd Textgraphs Workshop on Graph-Based Algorithms for Natural Language Processing, Manchester, United Kingdom, 2008.
\bibitem{chicoisne} Chicoisne, G., Blondin Mass\'e, A., Picard, O., Harnad, S. \emph{Grounding abstract word definitions in prior concrete experience}, 6th Int. Conf. on the Mental Lexicon, Banff, Alberta, 2008. 
\bibitem{cortese} Cortese, M. J., Khanna, M. M. \emph{Age of acquisition ratings for 3,000 monosyllabic words}, Behavior Research Methods, 40, 791--794, 2008.
\bibitem{harnad} Harnad, S. \emph{The symbol grounding problem}, Physica D 42:335-346, 1990
\bibitem{karp} Karp, R.M., \emph{Reducibility among combinatorial problems}. In R.E. Miller, J.W. Thatcher (Eds.), Complexity of Computer Computations, Plenum Press, New York, 1972, pp. 85--103.
\bibitem{porter} Porter M.F., \emph{An algorithm for suffix stripping}, Program, 14(3) 130--137, 1980.
\bibitem{ldoce} Procter, P., \emph{Longman Dictionary of Contemporary English (LDOCE)}. Longman Group Ltd., Essex, UK, 1978.
\bibitem{cide} Procter, P., \emph{Cambridge International Dictionary of English (CIDE)}. Cambridge University Press, 1995.
\bibitem{rosen} Rosen, K.H., \emph{Discrete mathematics and its applications}. 6th ed. McGraw-Hill, 2007.
\bibitem{stadthagen} Stadthagen-Gonzalez, H., and Davis, C. J. \emph{The Bristol norms for age of acquisition, imageability, and familiarity}, Behavior Research Methods, 38, 598--605, 2006.
\bibitem{tenenbaum} Steyvers, M. \& Tenenbaum, J.B., \emph{The large-scale structure of semantic networks: statistical analyses and a model of semantic growth}, Cognitive Science, 29(1), 41--78, 2005.
\bibitem{mrc} Wilson, M., Qx, O.O., Quinlan, P. \emph{MRC Psycholinguistic Database: Machine Usable Dictionary, version 2.00}, 1987.
\end{thebibliography}
